\documentclass[runningheads]{llncs}

\usepackage{eccv}

\usepackage{eccvabbrv}

\usepackage{graphicx}
\usepackage{float}
\usepackage{booktabs}
\usepackage{amsmath}
\usepackage{amssymb}
\usepackage{multirow}
\usepackage{pdfpages}
\IfFileExists{fontawesome5.sty}{%
  \IfFileExists{FontAwesome5Free-Solid.otf}{\usepackage{fontawesome5}}{}%
}{}

\graphicspath{{figures/}{./}}

\IfFileExists{axessibility.sty}{\usepackage[accsupp]{axessibility}}{} % accessibility aid; loads on Overleaf, skipped if absent

\usepackage[breaklinks,colorlinks,citecolor=eccvblue]{hyperref}

\begin{document}

\title{SVCBench: A Streaming Video Counting Benchmark for Spatial-Temporal State Maintenance}

\titlerunning{SVCBench}

\author{Pengyiang~Liu\inst{1,2} \and
Zhongyue~Shi\inst{1} \and
Hongye~Hao\inst{1} \and
Qi~Fu\inst{1} \and
Xueting~Bi\inst{1} \and
Siwei~Zhang\inst{1} \and
Xiaoyang~Hu\inst{1} \and
Zitian~Wang\inst{3}$^\dagger$ \and
Linjiang~Huang\inst{1} \and
Si~Liu\inst{1}}

\authorrunning{Liu et al.}

\institute{\fontsize{7.6pt}{9pt}\selectfont Institute of Artificial Intelligence, Beihang University, Beijing, China \and
Meituan, Beijing, China \and
Hangzhou International Innovation Institute of Beihang University, Hangzhou, China\\
\email{\{buaaplay, wangzt.kghl\}@gmail.com, \{ljhuang, liusi\}@buaa.edu.cn}}

\maketitle
\begingroup
\makeatletter
\renewcommand{\@makefnmark}{}
\footnotetext[1]{\textsuperscript{\textdagger}\,Corresponding author}
\makeatother
\endgroup

\vspace{-1.2em}
\begin{center}
{\small Project Page:~\href{https://buaa-colalab.github.io/SVCBench/}{\texttt{https://buaa-colalab.github.io/SVCBench}}}
\end{center}
\vspace{-0.5em}

\begin{abstract}
Video understanding requires models to continuously track and update world state during playback. Although existing benchmarks have advanced video understanding evaluation across multiple dimensions, they provide limited visibility into how models maintain world state over time. We propose SVCBench, a Streaming Video Counting Benchmark that repositions counting as a minimal, controlled probe for diagnosing models' world-state maintenance capability. We decompose this capability into object counting and event counting, forming 8 fine-grained subcategories. Object counting covers tracking currently visible objects and cumulative unique identities, while event counting covers detecting instantaneous actions and tracking complete activity cycles. SVCBench contains 406 videos with frame-by-frame annotations of 10,071 event occurrences and object state changes, yielding 1,000 streaming QA pairs with 4,576 query points distributed along video timelines. By observing state maintenance trajectories through streaming multi-point queries, we design three complementary metrics to diagnose numerical precision, trajectory consistency, and temporal awareness. Evaluations of mainstream video-language models show that current models still exhibit significant deficiencies in spatial-temporal state maintenance, with especially poor performance on periodic event counting. SVCBench provides a diagnostic framework for measuring and improving state maintenance in video understanding systems. Our code and data are available at \href{https://buaa-colalab.github.io/SVCBench/}{https://buaa-colalab.github.io/SVCBench}.
\end{abstract}

\section{Introduction}
\label{sec:intro}

\begin{figure}[t]
\centering
\includegraphics[width=\linewidth]{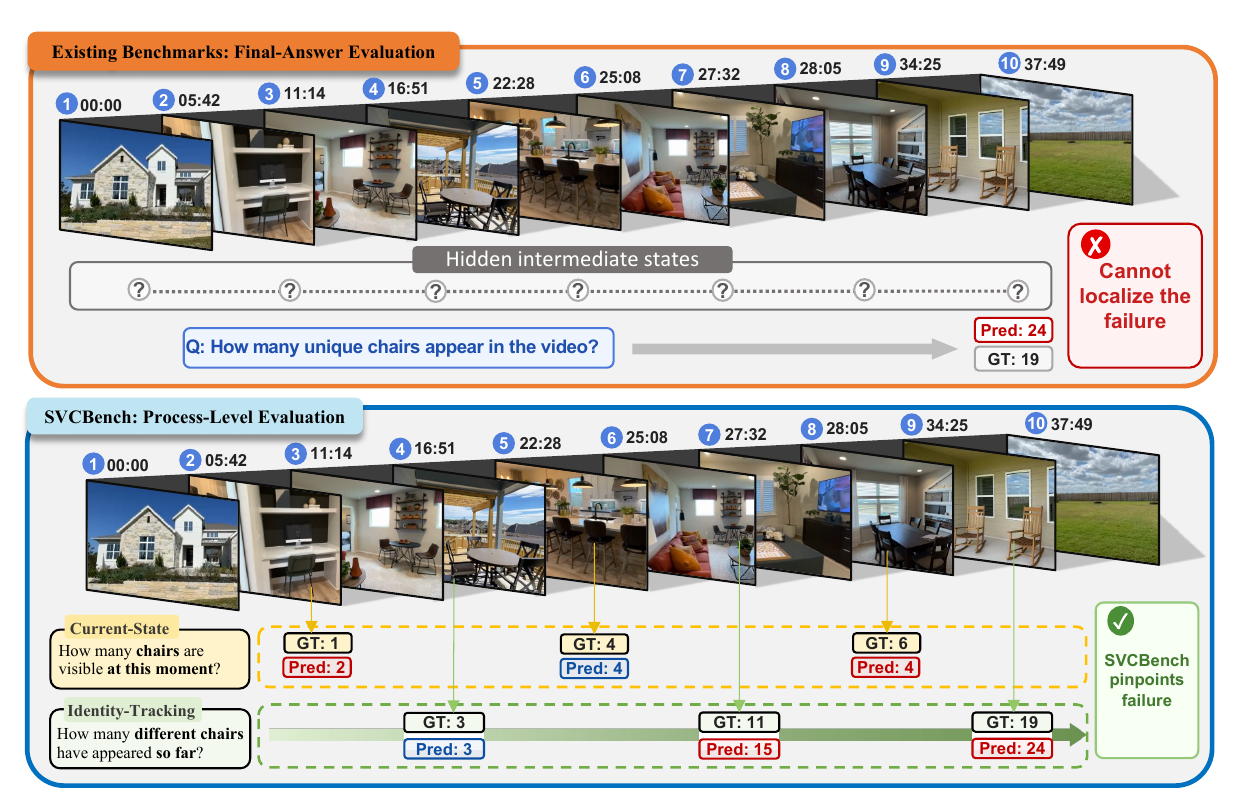}
\caption{Final-answer evaluation (top) only scores the last answer and cannot localize the error, whereas the process-level streaming queries in SVCBench (bottom) pinpoint when and where state maintenance fails.}
\label{fig:overview}
\end{figure}

Consider the scenario in Figure~\ref{fig:overview}. For the question ``How many chairs are visible at this moment?'', the answer changes as the camera moves. For the question ``How many different chairs have appeared so far?'', the answer should monotonically increase. These questions impose different requirements: the first requires tracking a current count that changes in real time; the second requires tracking cumulative unique identities. We refer to these internally tracked quantities, such as current counts, cumulative identities, and event occurrences, as \textbf{world state}, and the ability to continuously update them as \textbf{spatial-temporal state maintenance}. This ability is also closely related to embodied agents, which must continuously maintain and update task-relevant world state while perceiving and acting in dynamic environments. Throughout this paper, we use ``state maintenance'' as shorthand for spatial-temporal state maintenance. However, when models keep counts unchanged after objects leave, or produce decreasing counts in identity-tracking tasks, such contradictions expose defects in their state maintenance mechanisms.

\begin{figure}[t]
\centering
\includegraphics[width=\linewidth]{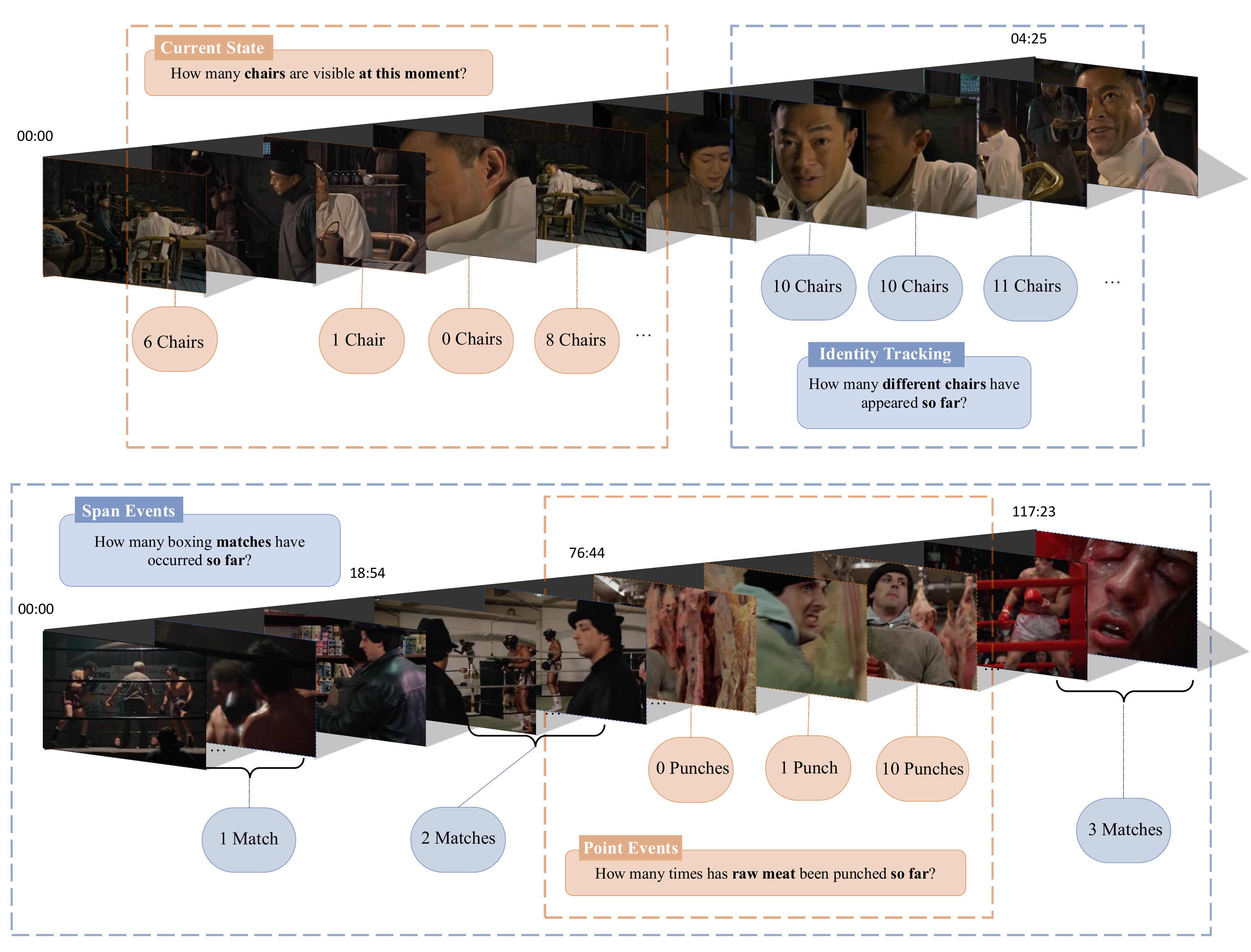}
\caption{Streaming counting as a probe for spatial-temporal state maintenance. Top: Object counting tracks the current count at each moment (current-state) and cumulative unique identities across time. Bottom: Event counting detects instantaneous actions and span events. Models are queried at multiple timepoints during video playback.}
\label{fig:teaser}
\end{figure}

Current video understanding benchmarks have made significant progress in standardization and capability coverage~\cite{fu2025video,shangguan2025tomato,yang2025thinking,niu2025ovo,wang2025lvbench,zhou2025mlvu,wu2024longvideobench,han2025videoespresso}, paralleling the rapid advances of vision-language models more broadly~\cite{niu2025native,niu2026mineru2,dong2026minerudiffusion}. However, we focus on an orthogonal dimension: \textbf{can existing evaluations observe how models maintain world state throughout video playback?} Most counting tasks present single queries where models provide a number after the video ends. Streaming benchmarks~\cite{lin2024streamingbench,niu2025ovo,xun2026rtv} introduce queries at multiple moments, but design choices---such as conflating tasks with different tracking requirements and using short temporal query windows---often prevent them from effectively evaluating models' spatial-temporal state maintenance.
To fill this gap, we propose SVCBench, a \textbf{S}treaming \textbf{V}ideo \textbf{C}ounting \textbf{Bench}mark for long videos. Our core design principle is to \textbf{reposition counting as a minimal probe for diagnosing how models maintain internal representations over time}. Specifically, counting questions must depend on well-defined quantities (such as the current object count in a scene, or the number of distinct individuals seen so far) and are difficult to bypass through vague semantic recognition or language priors. Numerical answers provide deterministic, option-interference-free supervision while avoiding the judgment dilemma of open-ended QA. We position counting as a controlled diagnostic entry point rather than a complete characterization of state maintenance: it yields objective, verifiable ground truth (e.g., ``14 chairs''), whereas many other states (e.g., ``has the person finished cooking?'') involve subjective boundaries or open-ended answers that are difficult to score reliably. More critically, by asking streaming questions at multiple timepoints during video playback, we can observe how models' predictions evolve rather than only checking isolated answers: if a model's prediction decreases from 5 to 3 in an identity tracking task, or fails to respond promptly to object entry/exit in a current-state task, such inconsistencies directly expose defects in its tracking mechanism. Based on this principle, we construct a systematic capability taxonomy that divides object counting into tracking currently visible objects and tracking cumulative unique identities, and event counting into detecting instantaneous actions and tracking complete activity cycles (Figure~\ref{fig:teaser}); these dimensions are further subdivided into eight fine-grained subcategories.

SVCBench contains 406 videos covering indoor navigation, first-person activities, sports videos, and other diverse scenes. We perform frame-by-frame annotation of 10,071 event occurrences and object state changes, and design 4,576 streaming query points across 1,000 QA pairs to observe how models' predictions evolve over time. We propose three complementary evaluation metrics to measure numerical precision, trajectory consistency, and temporal awareness. Systematic evaluation on mainstream video understanding models reveals significant deficiencies in current models' spatial-temporal state maintenance, providing clear directions for future model design.

\section{Related Work}
\label{sec:related}

\noindent\textbf{Long-Video Understanding Benchmarks.}
Video-MME~\cite{fu2025video} achieves a highly standardized comprehensive evaluation framework through multiple-choice questions, establishing full-spectrum coverage across multimodal tasks. LVBench~\cite{wang2025lvbench} pushes evaluation of long-term memory and understanding capabilities with extremely long videos (up to several hours). MLVU~\cite{zhou2025mlvu} systematically evaluates long video understanding through duration spans from 3 minutes to 2 hours with multi-task settings. LongVideoBench~\cite{wu2024longvideobench} focuses on referential reasoning in long contexts, where the core difficulty lies in retrieving referenced segments and completing reasoning. These benchmarks have advanced long video semantic understanding, but their evaluation signals are mostly task-level one-time answers, making it difficult to observe how models maintain world state along the timeline.

\noindent\textbf{Streaming and Online Video Evaluation.}
OVO-Bench~\cite{niu2025ovo}, as the most systematic online evaluation framework, proposes Backward Tracing, Real-Time Perception, and Forward Active Responding modes, evaluating models' dynamic reasoning and temporal awareness through precise timestamped queries at each timepoint. StreamingBench~\cite{lin2024streamingbench} focuses on understanding performance under real video stream input, covering real-time visual understanding, full-source information fusion, and contextual understanding subtasks. RTV-Bench~\cite{xun2026rtv} includes hundreds of hours of videos and thousands of QA samples, fine-grained testing models' responses to temporal information through extensive timestamp data. These benchmarks have advanced online video understanding evaluation, but their queries are often distributed within relatively short time windows, making it difficult to impose long-term cumulative state maintenance pressure on models, and existing online counting tasks often do not explicitly distinguish different tracking requirements for instantaneous actions versus extended activities.

\noindent\textbf{Counting and Temporal Reasoning Evaluation.}
TOMATO~\cite{shangguan2025tomato} constructs multi-task evaluation including action count, measuring temporal dependencies with metrics like multi-frame gain and sequence sensitivity. VSI-Bench~\cite{yang2025thinking} integrates counting into spatial intelligence evaluation, examining spatial relationships and object quantities under continuous observation of indoor scenes. CountLLM~\cite{yao2025countllm} focuses on repetitive action counting, emphasizing the role of periodic prompts and language supervision for counting generalization. AV-Reasoner~\cite{lu2025av} supports interpretable counting evaluation through clue annotation. These works have advanced counting and temporal reasoning evaluation from different perspectives, but most benchmarks emphasize temporal relationships within segments or specific counting subtasks, rarely making long-term cumulative state maintenance pressure and continuous updating of world state core design goals.

Complementary to existing work, SVCBench reformalizes counting as a minimal probe of world state, observing models' state maintenance trajectories through streaming multi-point queries, and explicitly aligning different types of state maintenance requirements to diagnosable sub-capabilities through a systematic capability decomposition framework.

\section{The SVCBench Benchmark}
\label{sec:method}

\subsection{Taxonomy: Decomposing State Maintenance via Counting}
\label{sec:taxonomy}

\begin{figure}[t]
\centering
\includegraphics[width=\linewidth]{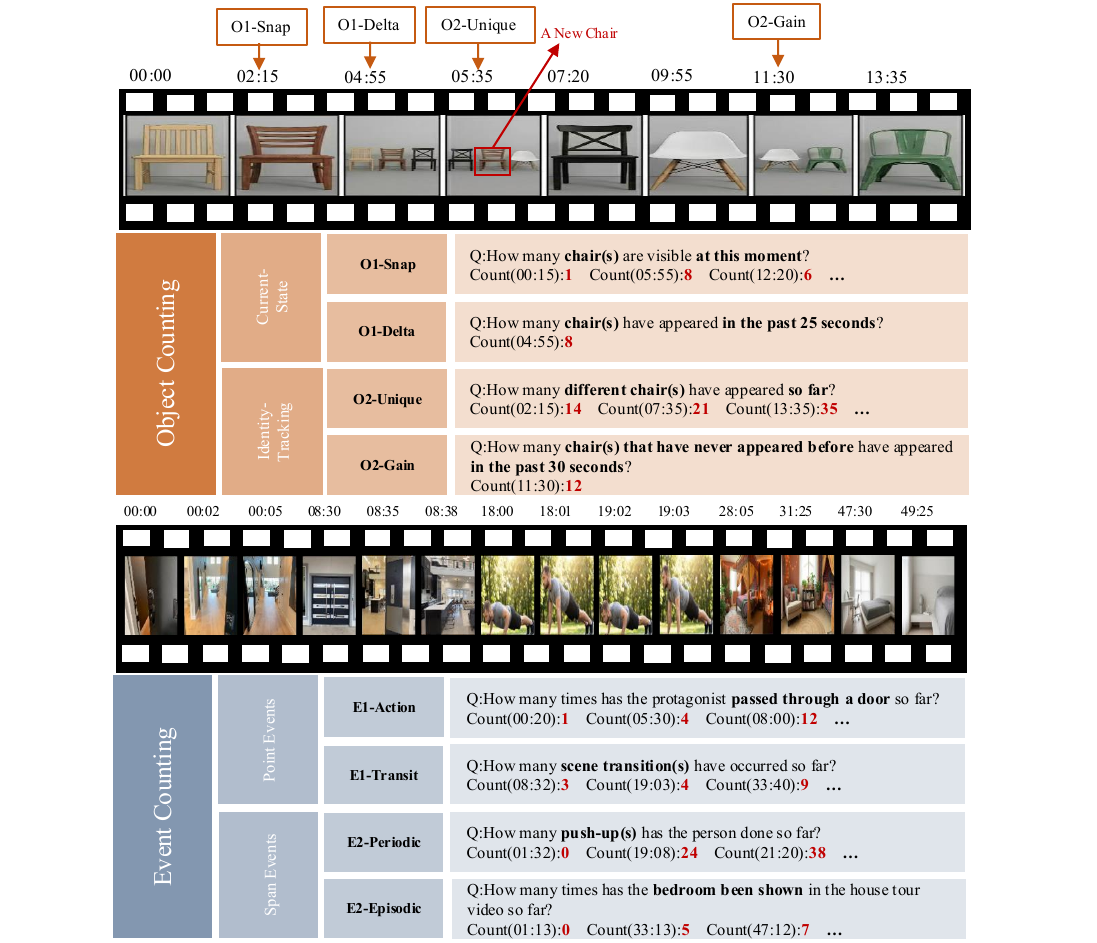}
\caption{Taxonomy of spatial-temporal state maintenance. We decompose counting into 8 subcategories across object and event dimensions. Each subcategory requires tracking different types of information: O1 tracks currently visible quantities, O2 tracks cumulative unique identities, E1 counts discrete instantaneous occurrences, and E2 counts complete activity cycles.}
\label{fig:taxonomy}
\end{figure}

We decompose spatial-temporal state maintenance capability into two orthogonal dimensions: object counting and event counting. Object counting focuses on tracking entities in the scene, while event counting focuses on identifying and tracking actions or activities in videos. Within each dimension, we subdivide tasks according to which quantities must be tracked and how those quantities update over time, forming the taxonomy shown in Figure~\ref{fig:taxonomy}.

\noindent\textbf{Object Counting.}
Object counting tracks entities in the scene. Based on query semantics, we divide it into current-state object counting (O1) and identity-tracking object counting (O2).

\noindent\textit{Current state (O1).}
O1 requires tracking the current count as it changes in real time. O1-Snap queries the number of objects visible in the scene at a given moment, so the model must update the count as objects enter or leave the view. O1-Delta queries the change relative to the start of a time window, which requires the model to retain the earlier count and compute the difference.

\noindent\textit{Cumulative identities (O2).}
O2 requires tracking cumulative unique identities across time. O2-Unique queries the total number of distinct individuals that have appeared across the timeline, so the model must judge whether each newly seen individual is new or a repeat. O2-Gain queries how many new individuals appear within a time window, combining identity tracking with a windowed constraint.

\noindent\textbf{Event Counting.}
Event counting identifies and tracks actions or activities over time. Based on temporal structure, we divide it into point event counting (E1) for instantaneous occurrences and span event counting (E2) for activities with temporal extent.

\noindent\textit{Point events (E1).}
E1 focuses on instantaneous atomic actions that can be approximated as a single point on the timeline. E1-Action queries the cumulative number of times a given atomic action occurs, where each occurrence is brief and acts as a discrete pulse. E1-Transit queries the number of scene transitions, where the model must detect the moment the scene switches from one stable configuration to another while maintaining the running total.

\noindent\textit{Span events (E2).}
E2 focuses on activities with clear start and end moments that occupy an interval on the timeline. E2-Periodic queries the number of complete, repetitive action cycles, which requires identifying the start and end boundaries of each cycle. E2-Episode queries the number of semantically complete activity segments, where the difficulty lies in boundary determination that depends on high-level semantic understanding.

O1 categories test models' ability to track current counts in real time, with prediction trajectories that can rise or fall; O2 categories test cross-temporal identity deduplication capability, with prediction trajectories that should be monotonically non-decreasing. E1 categories test precise moment detection capability, while E2 categories further require tracking complete event lifecycles. By observing models' prediction trajectories at multiple query points along the timeline, we can diagnose specific weaknesses in state initialization, incremental updates, and cross-temporal queries.

\subsection{Dataset Construction}
\label{sec:dataset}

\begin{figure}[t]
\centering
\includegraphics[width=\linewidth]{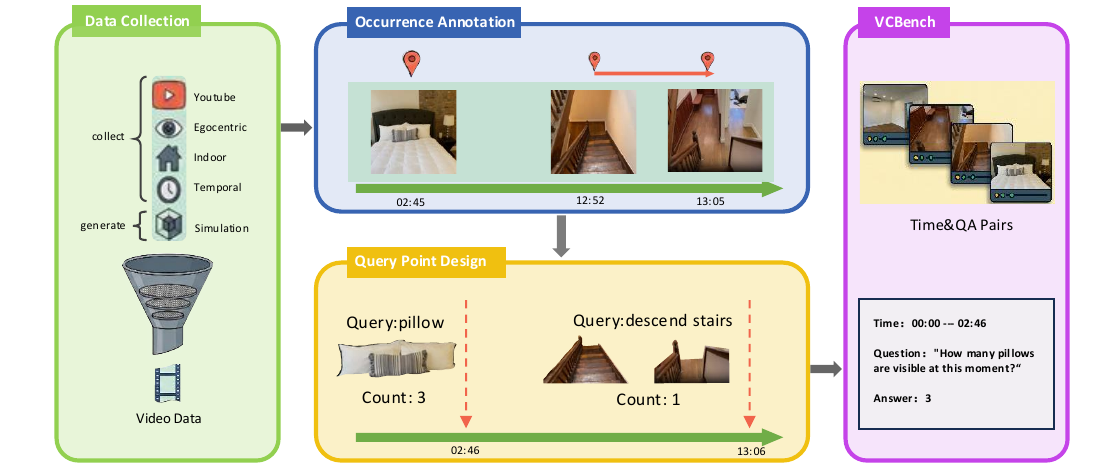}
\caption{SVCBench dataset construction pipeline. It shows the complete process from video source selection and fine-grained annotation generation to streaming query-point design. We collect videos from multiple sources including web platforms, existing computer vision datasets, and self-generated simulations, then perform manual annotations and design streaming query points.}
\label{fig:construction}
\end{figure}

Figure~\ref{fig:construction} outlines the construction pipeline, from video sourcing through frame-by-frame annotation to streaming query-point design.

\noindent\textbf{Video Sources and Special Processing.}
We collect videos from multiple sources: web platforms including YouTube (108 videos), existing computer vision datasets including ARKitScenes~\cite{baruch2021arkitscenes}, ScanNet~\cite{dai2017scannet}, ScanNet++~\cite{yeshwanth2023scannetpp}, Ego4D~\cite{grauman2022ego4d}, RoomTour3D~\cite{han2025roomtour3d}, TOMATO~\cite{shangguan2025tomato}, CODa~\cite{li2022coda}, and OmniWorld~\cite{zhou2025omniworld}, as well as self-generated physics-simulation videos. For the E2-Periodic category, we carefully selected 56 periodic action videos from TOMATO with seamless loop points and extended them through loop concatenation to generate 2--3 minute videos, addressing the scarcity of long-duration periodic video data.

\noindent\textbf{Annotation Protocol.}
The annotation process contains two core components. First, we perform frame-by-frame annotation of event occurrences and object state changes. Specifically, we annotate the timestamp of each occurrence for E1, the start and end moments for E2, and the first appearance moment of each individual for O2. Second, we design query points along the timeline, selecting moments with stable frames to avoid visual ambiguity. The correct answer for each query point is calculated based on annotated moments before that moment. O1-Delta and O2-Gain each set only 1 query point per question, as they focus on changes within specific time windows.

\begin{figure}[t]
\centering
\includegraphics[width=\linewidth]{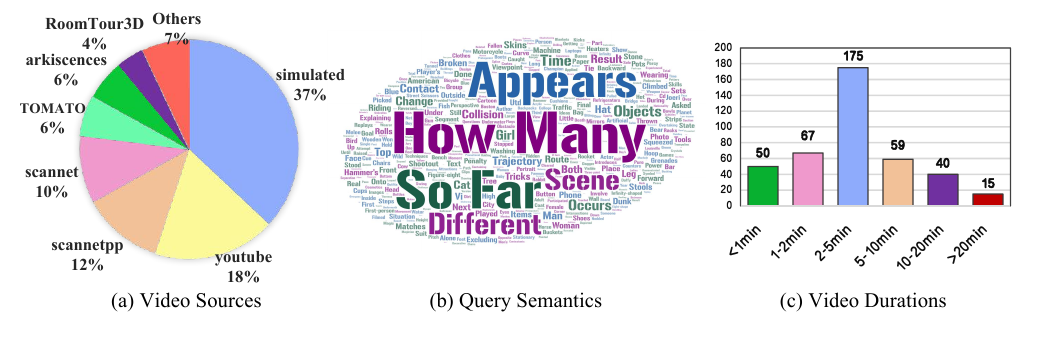}
\caption{SVCBench dataset analysis. (a) Pie chart of video source distribution, showing scene diversity across data sources. (b) Word cloud of query target semantics, showing the diversity of counted objects and events. (c) Histogram of video duration distribution.}
\label{fig:analysis}
\end{figure}

\noindent\textbf{Dataset Statistics.}
SVCBench contains 406 videos, 1,000 counting questions, 10,071 precisely annotated event occurrence moments and object state change moments, and 4,576 query points. Table~\ref{tab:dataset-stats} shows detailed statistics for each subcategory. Figure~\ref{fig:analysis} further illustrates the distribution of video sources, the diversity of counting-target semantics, and the distribution of video durations.

\begin{table}[t]
\centering
\small
\caption{Dataset statistics by subcategory. Annotated events are independent occurrences. E2 events include start and end timestamps. Avg.\ Density denotes the average number of queried targets per QA; ``-'' marks entries that are undefined for state-querying categories (O1-Snap/O1-Delta/O2-Gain), whose answers correspond to an instantaneous count or a window difference rather than enumerated events.}
\label{tab:dataset-stats}
\begin{tabular}{l|c|c|c|c}
\toprule
Category & Questions & Videos & Annotated Events & Avg. Density \\
\midrule
\textbf{Object Counting} & \textbf{546} & \textbf{196} & \textbf{2,483} & \textbf{4.5} \\
O1-Snap & 81 & 78 & - & - \\
O1-Delta & 98 & 75 & - & - \\
O2-Unique & 289 & 171 & 2,483 & 8.6 \\
O2-Gain & 78 & 64 & - & - \\
\midrule
\textbf{Event Counting} & \textbf{454} & \textbf{281} & \textbf{7,588} & \textbf{16.7} \\
E1-Action & 203 & 146 & 1,893 & 9.3 \\
E1-Transit & 48 & 29 & 682 & 14.2 \\
E2-Episode & 147 & 100 & 1,033 & 7.0 \\
E2-Periodic & 56 & 56 & 3,980 & 71.1 \\
\bottomrule
\end{tabular}
\end{table}

\subsection{Evaluation Metrics}
\label{sec:metrics}

SVCBench's streaming multi-point queries produce prediction trajectories rather than isolated answers. We design three complementary metrics to measure numerical precision, trajectory consistency, and temporal awareness.

\noindent\textbf{GPA (Gaussian Precision Accuracy)} measures the match between predictions and correct answers. For a question $q$ with $n$ query points, prediction sequence $P = [p_1, \ldots, p_n]$, and correct answer sequence $G = [g_1, \ldots, g_n]$:
\begin{equation}
\text{GPA}_q = \frac{1}{n} \sum_{i=1}^{n} \exp\left(-\frac{(p_i - g_i)^2}{2\sigma_i^2}\right), \quad \sigma_i = 0.05 \cdot \max(g_i, 1)
\end{equation}
By setting $\sigma$ as 5\% of the correct answer, the Gaussian kernel assigns near-zero scores to predictions with more than 15\% relative deviation ($3\sigma$), effectively penalizing significant outliers. GPA $\in [0, 1]$ applies to all 8 subcategories.

\noindent\textbf{MoC (Monotonicity Consistency)} verifies whether prediction sequences maintain the monotonic non-decreasing property for cumulative tasks. For a question $q$ with $n \geq 2$ query points, define the first violation position $v = \min\{i : p_{i+1} < p_i\}$ (if none exists, $v = n$):
\begin{equation}
\text{MoC}_q = \frac{v-1}{n-1}
\end{equation}
MoC $\in [0, 1]$, where 1 indicates a completely monotonic non-decreasing trajectory.

\noindent\textbf{UDA (Update Detection Accuracy)} measures whether models update their predictions at the moments when the world state actually changes. Let $S = \{i : g_i \neq g_{i-1}\}$ be the set of adjacent query points across which the ground-truth count changes. UDA is the fraction of these state-change steps at which the model also updates its prediction:
\begin{equation}
\text{UDA}_q = \frac{1}{|S|} \sum_{i \in S} \mathbb{1}[p_i \neq p_{i-1}]
\end{equation}
UDA $\in [0, 1]$ (undefined when $|S| = 0$), where 1 indicates the model responds at every state-change moment. MoC checks whether the trajectory regresses, while UDA checks whether the model updates exactly when the ground truth changes.

The three metrics form a complementary diagnostic framework: GPA captures numerical precision, MoC captures trajectory consistency, and UDA captures update timing. Their combination distinguishes failure modes. For instance, high GPA and MoC but low UDA indicates numerically accurate, monotonic predictions whose updates lag the true state changes, whereas low GPA together with low MoC indicates large numerical deviation and an unstable trajectory.

\section{Experiments}
\label{sec:experiments}

\subsection{Experimental Setup}

\noindent\textbf{Evaluated Models.}
We evaluate three types of systems to comprehensively compare spatial-temporal state maintenance capabilities under different architectural paradigms. \textbf{Offline multimodal models} include proprietary models (Gemini-3-Flash~\cite{google_gemini3_blog2025}, Doubao-Seed-1.8~\cite{doubao_seed18}, Kimi-K2.5~\cite{team2026kimi}, GPT-5.4~\cite{openai_gpt54_model_2026}) and open-source models (Qwen3-VL-8B/30B~\cite{bai2025qwen3}, Qwen2.5-VL-7B~\cite{qwen2025qwen25technicalreport}, InternVL-3.5-8B~\cite{wang2025internvl3}, Molmo2-8B~\cite{clark2026molmo2}, and the Mixture-of-Experts Qwen3.5-35B-A3B~\cite{qwen3_5}). These models independently process videos truncated to each query point. \textbf{Online video understanding models} include StreamingVLM~\cite{xu2025streamingvlm} (with a Qwen2.5-VL-7B backbone), Dispider~\cite{qian2025dispider}, LiveStar~\cite{livestar}, and Flash-VStream-7B~\cite{flashvstream}, which process video frames in a streaming manner, continuously updating internal state and responding to queries at any moment without reprocessing historical frames. We also introduce \textbf{GPT-4-Turbo}~\cite{achiam2023gpt} as a blind language model baseline, receiving only text questions without video frames, to quantify the actual contribution of visual information. \textbf{Human agents} complete evaluation under the same streaming query setting as models, watching videos and pausing to answer at each query point without rewatching played segments, providing capability upper bounds for model evaluation.

\noindent\textbf{Implementation Details.}
Offline models are evaluated using the protocol proposed by OVO-Bench~\cite{niu2025ovo}: for each query point, videos are truncated to that moment as input based on the model's video input limitations. For proprietary models with a fixed frame budget (Doubao-Seed-1.8, Kimi-K2.5, and GPT-5.4), we uniformly sample 64 frames from the truncated video at each query point, whereas API models that accept native frame rates (Gemini-3-Flash) are queried at 1\,fps up to their context limit. The query format is: ``Based on the video content up to this moment, [question] Please answer with a single number.'' Online models process video streams according to their native inference flow, injecting query questions upon reaching query moments and directly generating answers based on maintained internal state. All model outputs undergo unified post-processing to extract numbers, supporting both Arabic numerals and English number words; extraction failures are marked as invalid answers and scored as incorrect, contributing zero to GPA, so that a model is not rewarded for failing to produce a parseable count.

\subsection{Main Results}

Tables~\ref{tab:main_results_object} and~\ref{tab:main_results_event} show complete results for all evaluated models on SVCBench.

% Table 1: Overall + Object Counting
\begin{table*}[!t]
\centering
\scriptsize
\setlength{\tabcolsep}{2pt}
\caption{Overall Performance and Object Counting Results on SVCBench. GPA: Gaussian Precision Accuracy, MoC: Monotonicity Consistency, UDA: Update Detection Accuracy (scaled to 0--100). Best in \textbf{bold}, second \underline{underlined}.}
\label{tab:main_results_object}
\begin{tabular}{l|c|ccc|cc|c|ccc|c}
\toprule
\multirow{2}{*}{Model} & \multirow{2}{*}{Fr.} & \multicolumn{3}{c|}{Overall} & \multicolumn{2}{c|}{O1-Snap} & O1-Delta & \multicolumn{3}{c|}{O2-Unique} & O2-Gain \\
& & GPA & MoC & UDA & GPA & UDA & GPA & GPA & MoC & UDA & GPA \\
\midrule
\multicolumn{12}{l}{\textit{Human Performance}} \\
Human & - & 96.1 & 100.0 & 99.3 & 96.7 & 100.0 & 100.0 & 94.5 & 100.0 & 98.8 & 100.0 \\
\midrule
\multicolumn{12}{l}{\textit{Blind LLMs}} \\
GPT-4-Turbo~\cite{achiam2023gpt} & - & 18.7 & 95.7 & 4.3 & 15.7 & 0.6 & 19.4 & 15.8 & 96.4 & 3.6 & 50.0 \\
\midrule
\multicolumn{12}{l}{\textit{Proprietary Multimodal Models}} \\
Gemini-3-Flash~\cite{google_gemini3_blog2025} & 1fps & \textbf{37.0} & \underline{73.7} & \underline{73.8} & \textbf{44.3} & \underline{64.4} & \underline{41.0} & \textbf{36.5} & \textbf{77.8} & \textbf{79.8} & \underline{55.3} \\
Doubao-Seed-1.8~\cite{doubao_seed18} & 64 & \underline{36.2} & \textbf{77.2} & \textbf{76.8} & \underline{43.4} & \textbf{70.4} & \textbf{53.3} & \underline{32.5} & \underline{75.9} & \textbf{79.8} & \textbf{69.2} \\
Kimi-K2.5~\cite{team2026kimi} & 64 & 26.4 & 66.8 & 73.4 & 28.2 & 63.6 & 18.4 & 29.9 & 62.9 & \underline{78.2} & 24.4 \\
GPT-5.4~\cite{openai_gpt54_model_2026} & 64 & 29.1 & 71.8 & 59.3 & 26.6 & 45.5 & 26.5 & 30.9 & 72.2 & 57.1 & 35.9 \\
\midrule
\multicolumn{12}{l}{\textit{Open-Source Multimodal Models - Offline}} \\
Qwen3-VL-8B~\cite{bai2025qwen3} & 64 & \textbf{31.0} & \underline{84.3} & 55.1 & 21.1 & \textbf{43.8} & \textbf{35.7} & \underline{34.2} & \textbf{87.1} & 58.0 & \textbf{55.1} \\
Qwen3-VL-30B~\cite{bai2025qwen3} & 64 & 27.0 & \textbf{84.6} & \underline{56.4} & 19.1 & \underline{43.2} & 23.5 & 33.1 & 84.4 & \underline{63.9} & 26.9 \\
Qwen2.5-VL-7B~\cite{qwen2025qwen25technicalreport} & 64 & 19.1 & 68.2 & 40.8 & \textbf{29.7} & 38.3 & 20.4 & 23.8 & 67.1 & 53.2 & 46.2 \\
InternVL-3.5-8B~\cite{wang2025internvl3} & 64 & 24.2 & 81.6 & 55.6 & 20.3 & 37.4 & 6.1 & 30.5 & 82.2 & 57.8 & 37.2 \\
Molmo2-8B~\cite{clark2026molmo2} & 64 & 8.5 & 66.2 & 34.6 & \underline{26.5} & 42.6 & 1.0 & 13.0 & 53.4 & 46.8 & 3.8 \\
Qwen3.5-35B-A3B~\cite{qwen3_5} & 64 & \underline{29.3} & 82.4 & \textbf{59.0} & 20.9 & 41.6 & \underline{28.8} & \textbf{37.8} & \underline{85.6} & \textbf{70.8} & \underline{53.8} \\
\midrule
\multicolumn{12}{l}{\textit{Open-Source Multimodal Models - Online}} \\
StreamingVLM~\cite{xu2025streamingvlm} & 1fps & 19.1 & 68.1 & \textbf{50.3} & \textbf{36.7} & \textbf{60.3} & 6.1 & \textbf{26.0} & 65.6 & \underline{55.4} & 14.1 \\
Dispider~\cite{qian2025dispider} & 1fps & 7.7 & 34.8 & 15.5 & 11.4 & 8.2 & \underline{16.3} & 3.3 & 34.8 & 16.8 & 3.8 \\
LiveStar~\cite{livestar} & 1fps & \underline{19.9} & \textbf{86.4} & \underline{36.8} & \underline{24.1} & 16.0 & 13.3 & 20.9 & \textbf{81.8} & \textbf{57.3} & \underline{26.9} \\
Flash-VStream-7B~\cite{flashvstream} & 1fps & \textbf{23.4} & \underline{78.5} & 34.7 & 14.8 & \underline{36.6} & \textbf{45.9} & \underline{21.9} & \underline{80.8} & 37.9 & \textbf{56.4} \\
\bottomrule

\end{tabular}
\end{table*}

% Table 2: Event Counting
\begin{table*}[!t]
\centering
\scriptsize
\setlength{\tabcolsep}{2pt}
\caption{Event Counting Results on SVCBench. GPA: Gaussian Precision Accuracy, MoC: Monotonicity Consistency, UDA: Update Detection Accuracy (scaled to 0--100). Best in \textbf{bold}, second \underline{underlined}.}
\label{tab:main_results_event}
\begin{tabular}{l|ccc|ccc|ccc|ccc}
\toprule
\multirow{2}{*}{Model} & \multicolumn{3}{c|}{E1-Action} & \multicolumn{3}{c|}{E1-Transit} & \multicolumn{3}{c|}{E2-Episode} & \multicolumn{3}{c}{E2-Periodic} \\
& GPA & MoC & UDA & GPA & MoC & UDA & GPA & MoC & UDA & GPA & MoC & UDA \\
\midrule
\multicolumn{13}{l}{\textit{Human Performance}} \\
Human & 94.9 & 100.0 & 99.3 & 98.3 & 100.0 & 100.0 & 97.0 & 100.0 & 99.3 & 93.2 & 100.0 & 100.0 \\
\midrule
\multicolumn{13}{l}{\textit{Blind LLMs}} \\
GPT-4-Turbo~\cite{achiam2023gpt} & 13.1 & 96.5 & 4.2 & 23.6 & 95.8 & 4.2 & 21.9 & 93.7 & 7.1 & 0.0 & 94.2 & 5.4 \\
\midrule
\multicolumn{13}{l}{\textit{Proprietary Multimodal Models}} \\
Gemini-3-Flash~\cite{google_gemini3_blog2025} & \textbf{28.5} & 66.0 & 61.8 & \textbf{49.8} & 83.7 & \textbf{87.3} & \textbf{41.7} & \underline{79.5} & \underline{78.0} & \textbf{3.9} & \textbf{56.2} & \textbf{88.8} \\
Doubao-Seed-1.8~\cite{doubao_seed18} & 24.0 & \textbf{76.9} & \textbf{70.7} & 38.2 & \textbf{96.4} & 82.1 & \underline{40.5} & \textbf{84.4} & \textbf{79.3} & 0.8 & 50.4 & \underline{81.7} \\
Kimi-K2.5~\cite{team2026kimi} & 21.9 & 66.0 & \underline{70.3} & 33.5 & 74.8 & \underline{83.9} & 38.7 & 78.7 & 74.3 & 0.4 & \underline{52.2} & 63.4 \\
GPT-5.4~\cite{openai_gpt54_model_2026} & \underline{24.2} & \underline{71.5} & 58.5 & \underline{42.0} & \underline{85.9} & 82.1 & 37.5 & 79.1 & 70.6 & \underline{3.2} & 40.2 & 45.1 \\
\midrule
\multicolumn{13}{l}{\textit{Open-Source Multimodal Models - Offline}} \\
Qwen3-VL-8B~\cite{bai2025qwen3} & \textbf{22.5} & \underline{84.9} & \textbf{54.5} & \textbf{36.9} & \underline{95.3} & 70.8 & \textbf{35.6} & 82.1 & \underline{65.4} & 0.0 & \underline{68.8} & 29.0 \\
Qwen3-VL-30B~\cite{bai2025qwen3} & \textbf{22.5} & \textbf{88.1} & 46.4 & \underline{33.6} & 92.7 & 75.9 & \underline{35.3} & \textbf{89.0} & 64.5 & \textbf{2.5} & \underline{68.8} & \textbf{37.5} \\
Qwen2.5-VL-7B~\cite{qwen2025qwen25technicalreport} & 9.5 & 66.7 & 33.3 & 13.3 & 90.6 & 23.4 & 11.0 & 66.5 & 37.9 & 0.0 & 65.2 & \underline{32.1} \\
InternVL-3.5-8B~\cite{wang2025internvl3} & \textbf{22.5} & 80.2 & \underline{52.1} & 20.0 & 86.8 & \underline{77.1} & 31.8 & 81.3 & 65.3 & \underline{1.0} & 66.5 & 30.4 \\
Molmo2-8B~\cite{clark2026molmo2} & 3.5 & 71.7 & 21.7 & 2.7 & \textbf{95.5} & 7.6 & 9.4 & 73.7 & 35.3 & 0.4 & 66.1 & 30.4 \\
Qwen3.5-35B-A3B~\cite{qwen3_5} & \underline{19.0} & 78.1 & 47.6 & 27.7 & 87.2 & \textbf{80.6} & 29.3 & \underline{83.9} & \textbf{67.1} & 0.8 & \textbf{73.1} & 26.9 \\
\midrule
\multicolumn{13}{l}{\textit{Open-Source Multimodal Models - Online}} \\
StreamingVLM~\cite{xu2025streamingvlm} & 15.8 & 73.4 & \textbf{40.8} & \underline{13.8} & 58.5 & \textbf{61.3} & \underline{20.7} & 63.3 & \textbf{54.9} & \underline{0.4} & 82.6 & \textbf{24.1} \\
Dispider~\cite{qian2025dispider} & 11.8 & \underline{79.8} & 17.0 & 8.9 & 67.5 & 15.8 & 7.6 & \underline{84.0} & 14.7 & 0.0 & \textbf{96.4} & 4.9 \\
LiveStar~\cite{livestar} & \underline{16.1} & \textbf{89.5} & 23.5 & \textbf{22.8} & \textbf{91.7} & 31.9 & \textbf{23.7} & \textbf{87.1} & \underline{38.8} & \textbf{11.1} & \underline{93.2} & 8.9 \\
Flash-VStream-7B~\cite{flashvstream} & \textbf{19.0} & 75.5 & \underline{31.0} & 11.9 & \underline{84.2} & \underline{41.8} & 17.2 & 77.0 & 35.7 & 0.0 & 76.4 & \underline{19.9} \\
\bottomrule

\end{tabular}
\end{table*}

\noindent\textbf{Significant Human-Machine Gap.}
Human agents achieve GPA between 93--100 across all subcategories, while the best models (Gemini-3-Flash~\cite{google_gemini3_blog2025} and Doubao-Seed-1.8~\cite{doubao_seed18}) achieve overall GPA of only around 37, indicating spatial-temporal state maintenance is a broad bottleneck for current video understanding models. Notably, humans achieve GPA of 93.2 on E2-Periodic, whereas the best model on this subcategory reaches only 11.1 and almost all others fall below 4, highlighting the difficulty of periodic event counting.

\noindent\textbf{Overall Advantage of Proprietary Models.}
Proprietary models generally outperform open-source models in GPA (37 vs 27--31 for best open-source models), but open-source models show unexpected advantages in MoC (84+ vs 67--77), indicating open-source models tend to output smooth monotonic sequences but lack numerical precision. This may reflect different training strategies: proprietary models may be more aggressive in predicting large changes, but also more prone to violating monotonicity constraints. Greater scale or general capability does not by itself yield better state maintenance. GPT-5.4~\cite{openai_gpt54_model_2026} reaches only 29.1 overall GPA, below Gemini-3-Flash and Doubao-Seed-1.8, and the Mixture-of-Experts Qwen3.5-35B-A3B~\cite{qwen3_5} leads open-source models on update timing (overall UDA 59.0) but still lags in numerical precision (GPA 29.3).

\noindent\textbf{High MoC of Blind Baseline.}
GPT-4-Turbo~\cite{achiam2023gpt} achieves MoC of 95.7 without any visual input, indicating pure language models understand the language prior that cumulative counts should increase. However, its GPA is only 18.7 and its UDA is only 4.3, indicating visual information is crucial for accurate state maintenance and timing perception.

\noindent\textbf{Trade-offs Between Online and Offline Models.}
At comparable parameter scale, online models do not uniformly underperform offline ones in overall GPA. Flash-VStream-7B (7B) reaches 23.4, the strongest in overall GPA among the online models we evaluate and comparable to or better than 7--8B offline models such as Qwen2.5-VL-7B (19.1). The online disadvantage instead emerges over time: when an online model is compared against its own offline backbone under identical query sequences, accuracy degrades faster along the trajectory (Sec.~\ref{subsec:online-offline}), likely because they must compress historical information into fixed-size state representations. However, online models show higher MoC in some subcategories, indicating continuous internal state maintenance makes prediction trajectories smoother. Online models do not need to reprocess historical frames at each query point, which theoretically gives them computational advantages, but the cost is that state compression may lead to information loss.

\subsection{Controlled Experiments on State Maintenance Failures}

To deeply understand models' failure mechanisms in spatial-temporal state maintenance, we design two controlled experiments targeting cycle boundary detection in event counting and identity persistence tracking in object counting.

\subsubsection{Visual-Temporal Grounding via Explicit Count Annotation}
\label{subsec:oracle}

E2-Periodic (counting complete repetitive action cycles) presents the most severe failure mode we observe: almost all evaluated models achieve GPA below 4 (the strongest offline model, Gemini-3-Flash~\cite{google_gemini3_blog2025}, reaches only 3.9; even the best model overall, the online LiveStar~\cite{livestar}, attains just 11.1), while human agents achieve 93.2. Despite GPA approaching 0, models' MoC remains relatively high, indicating models are not randomly guessing but tend to output monotonically increasing sequences. This suggests models understand the language prior that cumulative counts should increase but largely fail to identify cycle boundaries from visual input.

To isolate visual perception capability from reasoning capability, we design an oracle experiment in which we evaluate Gemini-3-Flash~\cite{google_gemini3_blog2025} on 56 E2-Periodic questions from the TOMATO dataset~\cite{shangguan2025tomato}, comparing clean videos with videos overlaid with explicit cycle count annotations in the upper-left corner (e.g., ``Current count: 5''). This annotation directly tells the model the number of completed cycles, effectively bypassing the visual-temporal alignment bottleneck and directly testing the model's numerical reasoning capability.

\begin{figure}[t]
\centering
\begin{minipage}[t]{0.48\linewidth}
\centering
\includegraphics[width=\linewidth]{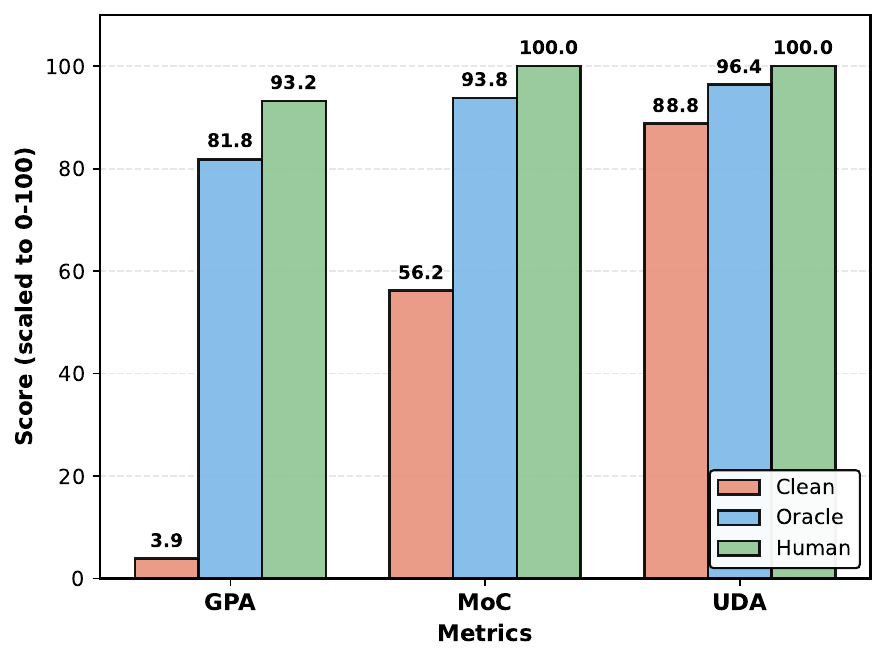}
\caption{Impact of explicit count annotation on E2-Periodic performance (scaled to 0--100).}
\label{fig:oracle}
\end{minipage}
\hfill
\begin{minipage}[t]{0.48\linewidth}
\centering
\includegraphics[width=\linewidth]{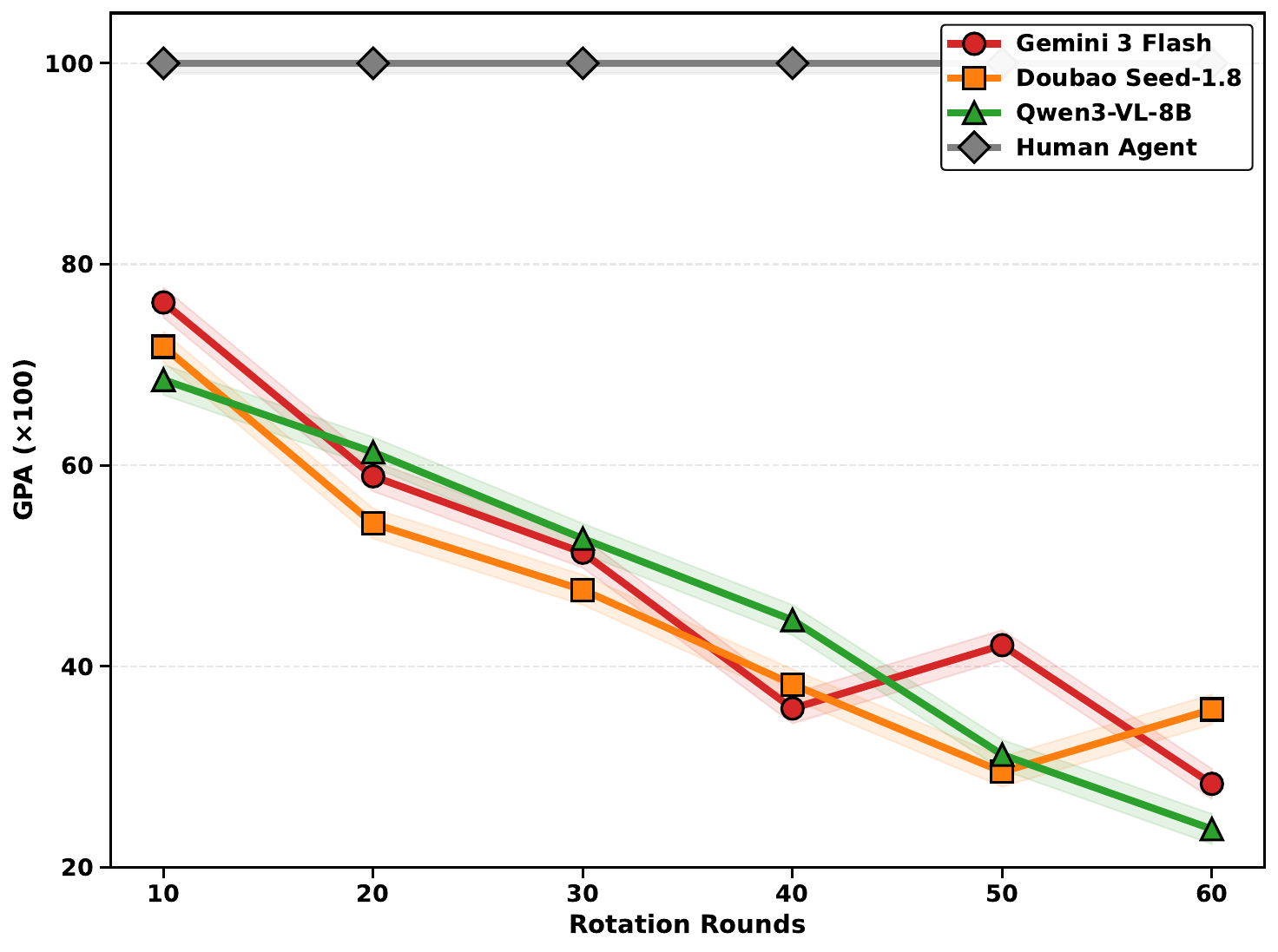}
\caption{Identity persistence degradation in rotating camera views (GPA scaled to 0--100).}
\label{fig:rotation}
\end{minipage}
\end{figure}

As Figure~\ref{fig:oracle} shows, explicit annotation improves GPA from 3.9 to 81.8 (21$\times$), approaching human level (93.2). MoC improves by 37.6 points, with monotonicity violations dropping from 43.8\% to only 6.2\%. This large improvement suggests that models possess sufficient reasoning capability to maintain cumulative counts, but largely lack the ability to detect cycle boundaries from pure visual input. In other words, the bottleneck is not understanding that counts should increase, but recognizing when they should be incremented. This indicates current video-language models lack robust visual-temporal grounding mechanisms and struggle to map repetitive visual motion patterns to discrete cycle boundaries.

\noindent\textbf{Natural long-period videos.}
To verify that this failure is not an artifact of loop concatenation, we additionally collect 50 natural long-period videos from YouTube (sports activities, 2--5\,min, with natural cycle variation). Model performance on these natural videos remains low and consistent with the loop-concatenated set (e.g., Gemini-3-Flash 3.9$\to$2.9, Doubao-Seed-1.8 0.8$\to$0.5 GPA), while humans stay high (93.2$\to$91.5), suggesting the periodic-counting bottleneck is not solely an artifact of loop concatenation.

\subsubsection{Identity Persistence Degradation in Rotating Camera Views}

To test models' persistence maintenance capability in object identity tracking, we construct a controlled scenario in which a camera rotates 360 degrees in place indoors, with only 1 bench remaining stationary. As the camera rotates, the bench continuously enters and leaves the field of view, but these are all repeated appearances of the same bench. We query ``How many different benches have appeared so far?'' The correct answer should remain 1 throughout.

We let the camera rotate continuously for 60 cycles and calculate the average GPA over all query points within each 10-cycle interval to observe the degradation trend of models' identity persistence maintenance capability over time. Figure~\ref{fig:rotation} shows the performance of three mainstream models.

In the first 10 cycles, the three models achieve GPA of 76.2 (Gemini-3-Flash~\cite{google_gemini3_blog2025}), 71.8 (Doubao-Seed-1.8~\cite{doubao_seed18}), and 68.5 (Qwen3-VL-8B~\cite{bai2025qwen3}). As rotation cycles increase, all three models' GPA shows a downward trend with notable fluctuations. By 60 cycles, the three models' GPA drops to 28.3, 35.7, and 23.8 respectively, approaching random guessing levels, while human agents maintain perfect performance of 100.

This degradation curve indicates that models struggle to maintain a robust cross-temporal identity-persistence representation of the set of seen objects. As the number of object reappearances increases, models' internal state representation gradually degrades, tending to misidentify each reappearing bench as a new object. This experiment echoes Section~\ref{subsec:oracle}: the former exposes failure in event boundary detection, the latter exposes failure in object identity maintenance, both suggesting current video-language models have limited representations for continuously updatable world state.

\section{Conclusion}
\label{sec:conclusion}

We propose SVCBench, a streaming counting benchmark for systematically measuring the spatial-temporal state maintenance capability of video understanding models. By reformalizing counting as a minimal probe of world state, we establish a systematic taxonomy that decomposes state maintenance capability into fine-grained subcategories across object and event dimensions. Our streaming multi-point query design enables evaluation to observe prediction trajectories rather than isolated predictions, thereby revealing failure modes that final-answer evaluation is unable to surface.

Systematic evaluation on mainstream models exposes clear limitations. The pronounced failure on periodic event counting suggests current architectures lack robust mechanisms for tracking temporal structure from visual input. Our oracle experiment provides mechanistic insight: the 21$\times$ improvement brought by explicit count annotation suggests models possess reasoning capability but lack visual-temporal grounding. Similarly, the rotating camera experiment indicates models struggle to maintain identity persistence, often misidentifying repeatedly appearing objects as new entities. Together, these findings point to limitations in current architectures: models have learned language priors but largely fail to ground them in world state maintenance.

SVCBench provides a diagnostic framework for advancing video understanding. Our taxonomy and metric system provide actionable improvement targets. The benchmark's streaming evaluation protocol and fine-grained subcategories enable researchers to isolate specific failure modes and measure progress in state maintenance capabilities. We hope SVCBench will support the development of video models with genuine temporal reasoning capabilities, advancing from pattern matching toward robust world state tracking.

\subsubsection*{\ackname}
This research is supported in part by the National Natural Science Foundation of
China (No.~62461160308, U23B2010, 62576024), the Beijing Natural Science
Foundation (No.~L231011), the Fundamental Research Funds for the Central
Universities (No.~501RCQD2025141003), and BeiHang GanWei Project
(No.~502GWXM2024141001). We also acknowledge the support of the Beijing Nova
Program.

\bibliographystyle{splncs04}
\bibliography{main}

% ---------------------------------------------------------------
% Supplementary Material
% ---------------------------------------------------------------
\clearpage
\begin{center}
{\Large\bfseries SVCBench: Supplementary Material}
\end{center}
\par
\vspace{1em}
\appendix
\renewcommand{\theHsection}{appendix.\arabic{section}}

\section{Dataset Construction Details}
\label{sec:sup-dataset}

\subsection{Annotation Protocol Overview}

The annotation process of SVCBench consists of two core components: first, frame-by-frame annotation of event occurrence moments and object state change moments; second, designing query points along the timeline. All annotations are completed by trained annotators.

\noindent\textbf{Annotator Training and Quality Control.}
Five annotators were trained and qualified on a held-out calibration set before formal annotation, with disputes adjudicated by a senior annotator. Annotators first learn the definitions and judgment criteria of the 8 subcategories, complete training on example videos, and pass consistency tests before formal annotation. Per-category inter-annotator agreement is high (Cohen's Kappa: O2-Unique = 0.82, E2-Periodic = 0.84, all other categories > 0.85). During the annotation process, experts regularly sample and review annotation results, and discuss and adjudicate controversial cases.

\subsection{Category-Specific Annotation Guidelines}

\noindent\textbf{Object Counting.}

\noindent\textit{O1-Snap} queries the number of visible objects in the scene at a specific moment. Annotators select moments with stable frames along the timeline as query points and count the currently visible target objects point by point. Judgment criterion: objects must be visible in the current frame; partially occluded but recognizable objects count as visible.

\noindent\textit{O1-Delta} queries the change in the number of currently visible objects relative to the time window starting point. Annotators first count the baseline at the window starting point, then count again at the window endpoint, and calculate the net change as Delta = current - baseline. Each question sets only 1 query point at the window endpoint.
\noindent\textit{O2-Unique} queries how many different individuals have appeared from the beginning of the video to the current moment. Annotators assign a unique ID to each individual appearing for the first time and record the first appearance time. Identity recognition is based on comprehensive judgment of visual features including appearance, color, and shape. The answer at a query point equals the total number of different individuals whose first appearance occurred before that moment, and the answer sequence should be monotonically non-decreasing.

\noindent\textit{O2-Gain} queries how many previously unseen different individuals newly appeared within a specific time window. Annotators maintain a global individual ID list and count the number of individuals whose first appearance time falls within the window. Each question sets only 1 query point at the window endpoint.

\noindent\textbf{Event Counting.}

\noindent\textit{E1-Action} counts the cumulative number of occurrences of a certain instantaneous action in the video. Annotators scan the video frame by frame and mark the timestamp of each action occurrence with frame-level precision. Judgment criterion: action duration should be less than 1 second, and must be completed to count.

\noindent\textit{E1-Transit} counts the number of scene switches or state transitions in the video. Annotators identify the moment of each hard cut or state mutation. Judgment criteria: only count hard cuts, as gradual transitions do not count; state transitions must be stable changes.

\noindent\textit{E2-Periodic} counts the number of complete action cycles with repetitive nature. Annotators identify the periodic pattern of actions and mark the start time and end time for each complete cycle. Judgment criterion: only count complete cycles from starting posture through action and return to starting posture; incomplete cycles do not count. The answer at a query point equals the number of complete cycles completed before that moment.

\noindent\textit{E2-Episode} counts the number of activity segments with semantic completeness. Annotators identify segment boundaries based on high-level semantic understanding rather than low-level visual features, and mark the start and end times for each segment. Judgment criterion: segments must have clear beginnings and endings, forming complete semantic units.
\subsection{Query Point Design}

Query point selection follows three principles. First, \textbf{frame stability}: select moments with stable frames and no motion blur. Second, \textbf{temporal coverage}: query points should cover the entire video to capture dynamic changes in count values. Third, \textbf{key moment priority}: set query points around event occurrences and moments when count values change.

O1-Delta and O2-Gain each set only 1 query point per question at the window endpoint. Other subcategories set 3--8 query points per question, with the specific number adjusted according to video length and event density.

\subsection{Special Data Processing}

\noindent\textbf{Loop Concatenation for E2-Periodic.}
Due to the scarcity of long-duration periodic video data, we carefully selected 56 periodic action videos from the TOMATO dataset and generated 2--3 minute long videos through precise loop boundary annotation and loop concatenation. Specifically, annotators first mark the start and end points of a complete loop cycle in the original video, ensuring seamless visual connection when jumping from the end point back to the start point. For cases where a cycle contains multiple sub-actions such as drawing 3 triangles consecutively, annotators also need to mark split points within the cycle to accurately record the cycle count. Based on these precise loop boundary annotations, we repeatedly concatenate a single cycle to generate long videos, ensuring strict timestamp alignment without cumulative errors. This design requires models to continuously track cycle boundaries over long time spans, imposing higher demands on state maintenance capability.
\section{Evaluation Protocol}
\label{sec:sup-eval}

\subsection{Offline Model Evaluation}

Offline multimodal models adopt the evaluation protocol proposed by OVO-Bench: for each query point, the video is truncated to that moment as input. We use the VLMEvalKit unified evaluation framework for assessment, which supports standardized inference for various video-language models.

\noindent\textbf{Video Truncation Strategy.}
According to the model's video input limitations in terms of number of frames or duration, videos are truncated from the beginning to the query moment. For videos exceeding the model's input limit, a uniform sampling strategy is used to extract key frames.

\noindent\textbf{Prompt Template.}
The query format is ``Based on the video content up to this moment, [question] Please answer with a single number.'', where [question] is the specific counting question.

\noindent\textbf{Output Post-processing.}
All model outputs undergo a unified post-processing pipeline to extract numerical answers, supporting both Arabic numerals such as ``5'' and English number words such as ``five''. Outputs that fail extraction are marked as invalid answers and scored as incorrect (contributing zero to GPA), so that failing to produce a parseable count is not rewarded.

\subsection{Online Model Evaluation}

Online video understanding models process video frames in a streaming manner, continuously updating internal state and directly generating answers at query moments without reprocessing historical frames. We reproduce online models based on open-source repositories.

\subsection{Human Evaluation}

Human agents complete evaluation under the same streaming query setting as models: watching videos and pausing to answer at each query point, without being allowed to rewatch played segments. This setting ensures fairness between human and model evaluation, providing an upper bound reference for capability.

Human evaluation used 3 trained raters under the streaming protocol, with each video completed by 1 rater, averaging ${\sim}$4\,min per QA (${\sim}$67 person-hours in total).
\subsection{Hardware Configuration and Efficiency}

All evaluations are conducted on an 8-card NVIDIA A100 (80GB) cluster. The inference time for offline models depends on video length and the number of query points, averaging 5--15 seconds per query point. Online models only require one forward pass to process the entire video, averaging 30--60 seconds per video, demonstrating significant computational efficiency advantages in long video evaluation.

\section{Metric Design Rationale}
\label{sec:sup-metrics}

\subsection{GPA Hyperparameter Selection}

Gaussian Precision Accuracy (GPA) uses a Gaussian kernel to measure the closeness between predicted values and correct answers, with the core hyperparameter being the standard deviation $\sigma_i = 0.05 \cdot \max(g_i, 1)$, \ie, set to 5\% of the correct answer.

\noindent\textbf{Design Motivation.}
We observe significant differences in failure modes across different models. Small models often output 0 as answers due to conservative strategies, while some large models such as Doubao-Seed-1.8 sometimes output values more than ten times the ground truth. If using Normalized Absolute Error (NAE), the latter would be severely penalized, even though its counting is quite accurate in most cases.

The soft tolerance mechanism of the Gaussian kernel can more reasonably evaluate model performance: through the 3$\sigma$ principle, predictions with relative deviations within 15\% receive high scores, while predictions exceeding 15\% rapidly decay to near zero. This design maintains precision requirements while avoiding extreme outliers dominating the overall score.

\noindent\textbf{Robustness to $\sigma$.}
Model rankings are insensitive to the tolerance $\sigma$. Sweeping $\sigma \in \{0.05, 0.10, 0.15, 0.20\}\cdot\max(g_i,1)$, overall GPA increases smoothly with the tolerance (e.g., Gemini-3-Flash from 37.0 at $\sigma{=}0.05$ to 43.2 at $\sigma{=}0.20$; Doubao-Seed-1.8 from 36.2 to 42.3; StreamingVLM from 19.1 to 21.0), while the relative ordering is essentially unchanged (Kendall's $\tau > 0.95$ between any two settings). This indicates the low absolute scores reflect genuine model limitations rather than metric strictness.
\subsection{Complementary Diagnostic Value}

The three metrics are complementary. GPA measures numerical precision, i.e., the closeness between predicted values and correct answers. Monotonicity Consistency (MoC) measures trajectory consistency, examining monotonicity constraints for cumulative tasks. Update Detection Accuracy (UDA) measures update timing, examining whether the model responds when the world state changes.

Through joint analysis, different failure modes can be distinguished. High GPA combined with high MoC but low UDA indicates numerical accuracy and monotonicity but temporal perception failure, meaning the model updates counts at wrong moments. Low GPA combined with high MoC indicates large numerical deviations but smooth prediction trajectories, possibly over-relying on language priors. Low GPA combined with low MoC indicates large numerical deviations and unstable prediction trajectories, suggesting state maintenance mechanism collapse.

\section{Additional Robustness Analyses}
\label{sec:sup-robust}

\subsection{Statistical Reliability}
To confirm that the reported gaps are not an artifact of sample size, we compute bootstrap 95\% confidence intervals (1{,}000 resamples) on the per-model mean GPA. Even for the smallest subcategories the intervals are narrow, with E1-Transit at $[1.8, 5.2]$ and E2-Periodic at $[1.5, 4.6]$, far narrower than the human--model gaps, indicating the conclusions are statistically stable.

\subsection{Robustness to Sampling Rate}
We re-evaluate Qwen2.5-VL-7B at 0.5/1/2/4\,fps. Performance is stable and the trends are unchanged: E1-Action GPA 9.3/9.8/10.2/10.2 and E2-Periodic GPA 0.0/0.0/0.1/0.1 across the four rates, indicating our findings are robust to the frame-sampling choice.

\subsection{Controlled Online vs.\ Offline Comparison}
\label{subsec:online-offline}
To isolate the effect of streaming compression from model capacity, we compare StreamingVLM against its own offline backbone Qwen2.5-VL-7B under identical query sequences. Splitting each trajectory at its midpoint, the online model's GPA drops from 24.5 to 13.5 across the two halves, a decline about 2.5$\times$ steeper than the offline backbone (21.5$\to$17.0). Since both share the same backbone, the steeper online decline reflects progressive context loss under streaming state compression rather than a difference in base capability.

\subsection{Temporal Degradation Within a Video}
Splitting trajectories into early/mid/late thirds, Gemini-3-Flash's O2-Unique GPA declines monotonically (42.3$\to$35.8$\to$29.1) while its O1-Snap GPA stays stable (45.0/44.2/43.5). This indicates degradation stems from cumulative state pressure specific to identity tracking, not a general attention decay over long inputs.

\subsection{Prompt Format}
Chain-of-thought prompting yields near-identical performance ($\Delta$GPA $<$ 1) on representative models (Gemini-3-Flash and Qwen2.5-VL-7B), suggesting the observed failures are not an artifact of prompt format.

\section{Data Release and Licensing}
\label{sec:sup-license}
SVCBench reuses publicly released or publicly accessible videos and contributes frame-by-frame human annotations (event and state-change timestamps, per-individual first-appearance times, and streaming query points) together with the full evaluation code. The annotations and evaluation code are released under the CC BY 4.0 license, and our self-generated physics-simulation videos are released directly under the same license. The source videos retain their respective original licenses, and users must comply with the terms of each source: Ego4D (Ego4D License), ScanNet and ScanNet++ (their respective Terms of Use, non-commercial research), ARKitScenes (Apple ARKitScenes license), RoomTour3D (CC BY-NC), CODa (Apache-2.0), OmniWorld (CC BY-NC), TOMATO (academic, non-commercial use only: its self-created videos under CC BY-NC-SA, and its YouTube-sourced videos under the original creators' Creative Commons licenses), and other YouTube content (platform and creator terms). By downloading SVCBench, users agree to adhere to CC BY 4.0 for our annotations and code and to the licenses of all source datasets for the videos. The E2-Periodic clips are produced by loop-concatenating TOMATO videos; as derivatives of TOMATO they are provided for non-commercial academic research only and, in line with TOMATO's CC BY-NC-SA terms, are made available under the same CC BY-NC-SA license with attribution to TOMATO and to the original video creators. We additionally provide the loop-boundary annotations and the corresponding TOMATO source IDs. If you are a rights holder with any concern about specific content, please contact us and we will address it promptly.

\section{Per-Category Question Examples}
\label{sec:sup-examples}

This section presents representative question examples for the 8 subcategories through visualizations. Each page shows one subcategory with question text, key frames, ground truth answers, and typical model predictions.

\includepdf[pages=-,pagecommand={\thispagestyle{headings}}]{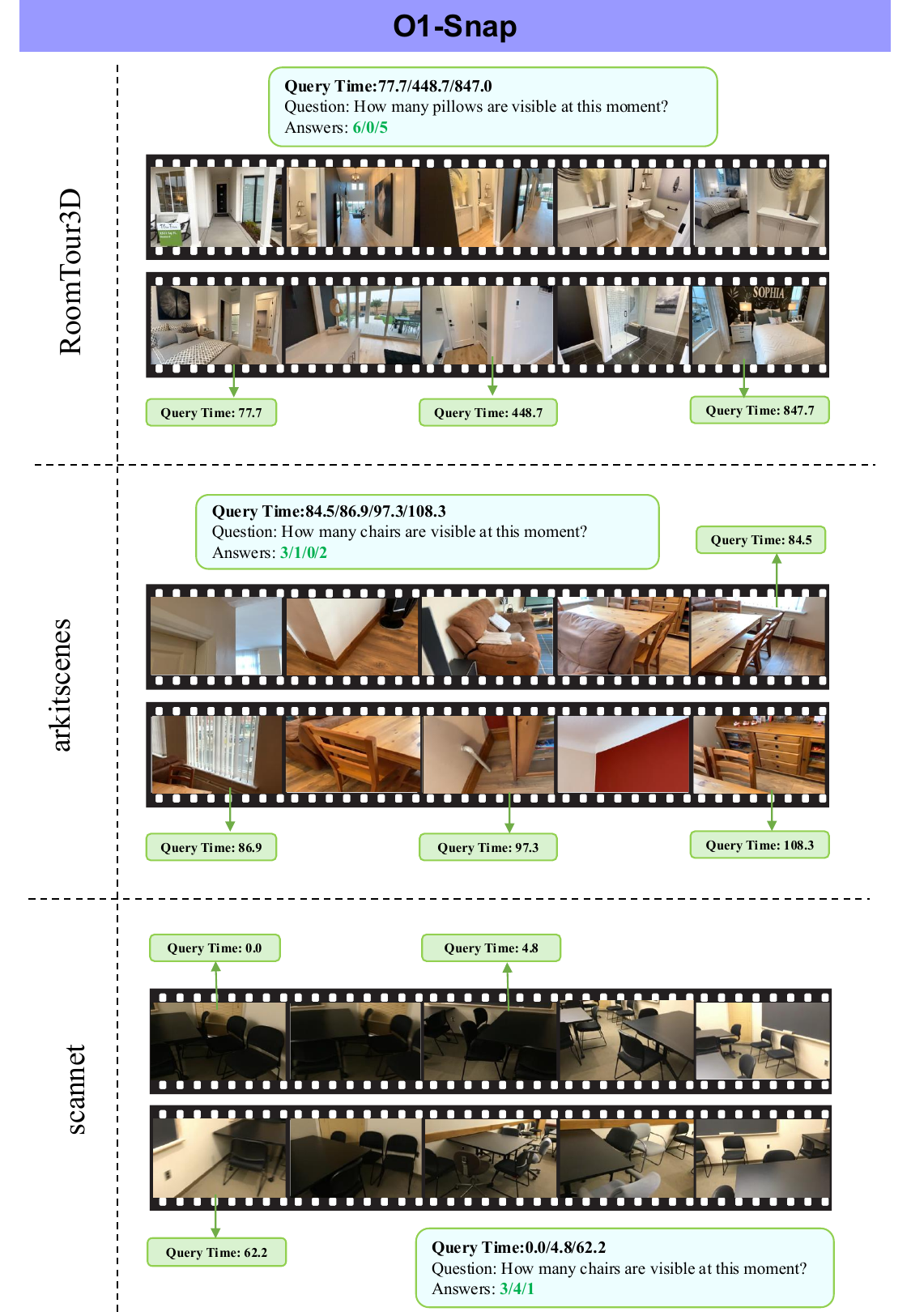}
\section{Limitations and Future Work}
\label{sec:sup-limitations}

\subsection{Dataset Limitations}

\noindent\textbf{Scale and Coverage.}
SVCBench contains 406 videos and 1,000 QA pairs. Due to the need for manual frame-by-frame annotation over long time spans in long videos, such as annotating complete cycle boundaries for E2-Periodic and tracking the first appearance moment of each individual for O2-Unique, annotation costs are high, resulting in a smaller scale compared to large-scale video understanding benchmarks. Future work can explore semi-automatic annotation methods, using object detection and action recognition models to assist annotators, expanding dataset scale while ensuring quality, and covering more scene types such as medical, industrial, and natural scenes, as well as more counting targets.

\subsection{Evaluation Protocol Limitations}

\noindent\textbf{Offline Evaluation Approximation.}
The offline evaluation protocol approximates streaming evaluation through video truncation, which may introduce discrepancies with true streaming processing. Specifically, the model sees the complete video from the beginning to the query point each time, rather than incrementally processing new frames. This may overestimate model performance because the model can re-examine historical content. Future work should develop more native streaming evaluation protocols.

\noindent\textbf{Online Model Coverage.}
We evaluate four online models (StreamingVLM, Dispider, LiveStar, and Flash-VStream-7B), yet none fully demonstrates the potential advantages of online architecture in state maintenance. This may be because current online models' state compression mechanisms are not sufficiently optimized, or the evaluation protocol fails to effectively distinguish the essential differences between the two architectures. Future work requires deeper analysis and evaluation of more online models to comprehensively understand the advantages and disadvantages of online versus offline architectures in state maintenance tasks.
\subsection{Metric Design Limitations}

\noindent\textbf{GPA Sensitivity to Large Values.}
When the correct answer is very large such as greater than 100, a 5\% relative tolerance may still correspond to a large absolute error. Future work can explore adaptive tolerance designs that dynamically adjust according to counting range.

\noindent\textbf{MoC Misleading in Extreme Cases.}
When model performance is very poor, MoC may produce misleading scores. For example, when a model outputs all 0s, MoC still achieves a perfect score because the sequence is monotonically non-decreasing, but this obviously does not represent good state maintenance capability. Future work should design more robust monotonicity metrics that can distinguish between trivial monotonicity and meaningful state tracking.

\subsection{Future Directions}

\noindent\textbf{Multimodal Input.}
Currently only visual input is used. Future work can introduce audio signals, such as counting knock times or number of speakers through sound, to evaluate models' multimodal state maintenance capability.

\noindent\textbf{Interactive Evaluation.}
Currently a passive observation mode. Future work can design interactive evaluation, allowing models to actively query key frames or request replay of segments, closer to real application scenarios.

\end{document}